\documentclass[11pt,a4paper]{article}
\usepackage[hyperref]{acl2019}
\usepackage{times}
\usepackage{latexsym}

\usepackage{url}

\usepackage{xspace}
\usepackage{url}
\usepackage{color}
\usepackage{xcolor}
\usepackage{multirow}
\usepackage{amsmath}
\usepackage{graphicx}
\usepackage{amsfonts}

\DeclareMathOperator*{\enc}{enc}
\DeclareMathOperator*{\dec}{dec}
\DeclareMathOperator*{\dif}{dif}

\DeclareMathOperator*{\bleu}{BLEU}
\DeclareMathOperator*{\maxbleu}{Max-BLEU}
\DeclareMathOperator*{\rouge}{ROUGE}
\DeclareMathOperator*{\maxrouge}{Max-ROUGE}

 \aclfinalcopy 

\title{Generating Diverse Story Continuations with Controllable Semantics}

\author{Lifu Tu$^1$ \ \ \ Xiaoan Ding$^2$ \ \ \  Dong Yu$^3$ \ \ \ Kevin Gimpel$^1$ \\[1ex]
$^1$Toyota Technological Institute at Chicago, Chicago, IL, 60637, USA\\
$^2$University of Chicago, Chicago, IL, 60637, USA\\
$^3$Tencent AI Lab, Bellevue, WA, 98004, USA\\
[1ex]
{\small
{\tt \{lifu, kgimpel\}@ttic.edu}, \ \ \
{\tt xiaoanding@uchicago.edu}, \ \ \
{\tt dyu@tencent.com}
}\\}

\date{}

\begin{document}
\maketitle
\begin{abstract}
We propose a simple and effective modeling framework for controlled generation of multiple, diverse outputs. 
We focus on the setting of generating the next sentence of a story given its context. 
As controllable dimensions, we consider several sentence attributes, including sentiment, length, predicates, frames, and automatically-induced clusters. 
Our empirical results demonstrate: (1) our framework is accurate in terms of generating outputs that match the target control values;
(2) our model yields increased maximum metric scores compared to standard $n$-best list generation via beam search; 
(3) controlling generation with semantic frames leads to a stronger combination of diversity and quality than other control variables as measured by automatic metrics. We also conduct a human evaluation to assess the utility of providing multiple suggestions for creative writing, demonstrating promising results for the potential of controllable, diverse generation in a collaborative writing system. 
\end{abstract}

\section{Introduction}
\label{intro}

We consider the problem of automatic story continuation generation, i.e., how to generate story continuations conditioned on the story context.
Inspired by recent work in controllable generation \citep{DBLP:conf/icml/HuYLSX17,DBLP:journals/corr/FiclerG17aa}, we propose a simple and effective modeling framework for controlled generation of multiple, diverse outputs based on interpretable control variables.
Each control variable corresponds to an attribute of a sentence. Compared to previous work that only seeks to control the values of sentiment \citep{DBLP:conf/icml/HuYLSX17} and length \citep{kikuchi-EtAl:2016:EMNLP2016}, 
we further explore neural text generation with particular verbal predicates, 
semantic frames, and automatically-induced clusters. 

We compare the diversity of story continuations controlled by different sentence attributes and find that using frames yields 
a stronger combination of diversity and quality than other control variables as measured by automatic metrics.
Unlike certain other attributes, frames have hundreds of possible values. Some frame values can help to get a natural continuation, while others are not applicable when considering the story context.

\begin{table}[t]
\small
\centering
\begin{tabular}{|ll|}
	\hline
	\textbf{Context:} & \\ \hline
	\multicolumn{2}{|p{0.9\columnwidth}|}{
    sandra needed a new phone . her old one had broken . she walked into the store and bought a new one . she was very excited .}\\
	\hline
	\multicolumn{2}{|l|}{\textbf{Control Attributes and Generated Continuations:}} \\ \hline
	\multicolumn{2}{|l|}{\textbf{Sentiment}:} \\ 
	\multicolumn{2}{|l|}{positive \quad\quad sandra was happy to have a new phone .} \\
	\hline
	\multicolumn{2}{|l|}{\textbf{Predicates}:} \\
	loved & sandra loved the new one . \\
	gave & sandra 's mom gave sandra a refund . \\
	\hline
   \textbf{Frames}:&\\
   Calendric\_unit & it was the perfect day of her life .\\ 
	Cardinal\_numbers & it was a perfect one . \\
	Activity\_ongoing & she kept it on the couch . \\
    \hline
    \end{tabular}
\caption{Story continuations conditioned on various control attributes generated from our framework.}
\label{tab:task}
\end{table}

We quantitatively evaluate both controllability and diversity. 
Our empirical results show that: (1) our framework is accurate in terms of generating outputs that match the target attribute values; (2) our framework increases maximum metrics scores compared to $n$-best list generation with beam search; (3) controlling generation with frames leads to a stronger combination of diversity and quality than other control variables as measured by automatic metrics. 
 
We also explore methods to rerank continuations to choose attribute values automatically and obtain a small number of high-quality outputs. 
We consider two frame ranking methods: one reranks the generated continuations using a reverse scoring model~\citep{LiGBGD15} and returns the $k$-best generations; the second first predicts the $k$ most likely frames based on the context, and uses these frames to generate continuations. One potential use case of controllable, diverse story generation is collaborative writing applications~\citep{ClarkRTJS18}. 
 
We conduct a human evaluation to assess the utility of providing multiple suggestions from our models in this
setting, demonstrating promising results for the potential of controllable generation for collaborative writing.

\section{Task Description and Definitions}
Given a story context and a control attribute value, our goal is to generate a story continuation that: (1) conforms to the given attribute value, (2) is relevant to the 
story context, and (3) is complementary to continuations with other control attribute values, thereby providing a diverse set of continuations when used with multiple attribute values. 

We use $x = \langle x_1, x_2,..., x_{|x|}\rangle$ to denote a story context and $y = \langle y_1, y_2,..., y_{|y|}\rangle$ to denote a story continuation. The last token $y_{|y|}$ is assumed to be $\langle \mathit{eos}\rangle$. We develop a framework to model tuples $(x, l, y)$, where $l$ is a control attribute. 
The control attribute represents a characteristic of the continuation, such as its length, sentiment, automatically-induced cluster, verbal predicate, or set of semantic frames. Table \ref{tab:task} shows several examples of control attributes and generated continuations corresponding to them from our model.

\section{Model}

Our controllable generation framework is a variation of a sequence-to-sequence (seq2seq) model with attention~\cite{NIPS2014_5346,BahdanauCB14,LuongPM15}. 
To represent the control attribute values, we define an attribute embedding function $R$ that maps a given attribute value $l$ to a vector $z$: $z = R(l)$. 
Here $l$ can be a single discrete number or a set of discrete numbers, depending on what attributes are being used. The control variable $z$ contains two parts: $z = [z_{\enc}; z_{\dec}]$ 
where semicolon (;) denotes vertical concatenation and  $z_{\enc}$  and $z_{\dec}$ are additional inputs for the encoder and decoder respectively.

\paragraph{Encoder.} For our encoder, we use a standard bidirectional recurrent neural network (RNN):
\begin{align*}
\overrightarrow{s_{i}}&=f_{e1}([v_{i}; z_{\enc}], \overrightarrow{s_{i-1}}) \\
\overleftarrow{s_{i}}&=f_{e2}([v_{i}; z_{\enc}], \overleftarrow{s_{i+1}}) \\
s_{i} &= [\overrightarrow{s_{i}};\overleftarrow{s_{i}}]
\end{align*}
where $v_i$ is the vector representation of word $x_i$, $s_{i} \in \mathbb{R}^d$ is the hidden state at time $i$, and $f_{e1}$ and $f_{e2}$ are the forward and backward RNN functions.

\paragraph{Decoder.} Our decoder uses an RNN with the general global attention scheme from \citet{LuongPM15}. 
An additional input $z_{\dec}$ is fed to the decoder at each time step to reflect the characteristics of the control variable: 
\begin{align}
h_j=f_d([y_{j-1}; z_{\dec}], h_{j-1}) \nonumber
\end{align}
\noindent where $h_j$ is the decoder RNN hidden vector at time step $j$ and $f_d$ is the decoder RNN function. Then, the conditional probability with controllable generation can be decomposed as follows:
\begin{align}
\log p_{\Theta}(y \mid x,l) = \sum_{j=1}^{|y|} \log p_{\Theta}(y_j \mid y_{<j},s, l)\nonumber
\end{align}
Here $s$ represents the hidden states of the source sequence and $\Theta$ are the parameters of the seq2seq attention model.

\paragraph{Training.}

Our training objective is:
\begin{align}
\min_{\Theta,  R} \sum_{\langle x, l, y\rangle \in \mathcal{D}}-\log p_{\Theta}(y \mid x, l)\label{eq:updateloss}
\end{align}
\noindent where $\mathcal{D}$ is the training data, i.e., we assume attribute values $l$ are provided during training. In practice, these will be predicted automatically using a linguistic analyzer. 

With certain attributes, we do not update the attribute value embeddings $R$, i.e., we fix $z_{\enc}$ and $z_{\dec}$ to one-hot vectors. 

\paragraph{Inference.} We can specify the value of the control variable and generate specific outputs. By changing the variable values, we obtain multiple continuations for the same context. Beam search is used for decoding.

\section{Control Attributes}

\begin{table}[t!]
\setlength{\tabcolsep}{3pt}
\small
\centering
\begin{tabular}{|p{0.965\columnwidth}|}
	\hline
	\textbf{Context:}
    i bought a pair of earbuds at target . i spent ten dollars . someone told me they were cheaper at the dollar store . they were only a dollar .\\
    \textbf{Gold Continuation}: i wish i had known that before . \\
    \hline 
\end{tabular}
\begin{tabular}{|p{0.18\columnwidth}p{0.76\columnwidth}|}
    \hline
    \multicolumn{2}{|l|}{\textbf{Sentiment}} \\
    \textbf{negative} & \textbf{i was disappointed .} \\
    neutral & i decided to keep them . \\
    positive & i was able to get a new pair . \\
    \hline
\end{tabular}
\begin{tabular}{|p{0.06\columnwidth}p{0.88\columnwidth}|}
    \hline
    \multicolumn{2}{|l|}{\textbf{Length}} \\
    4 &  i was shocked .\\
    5 &  i was rather disappointed .\\
    6 &  i bought a new pair .\\
    7 &  i was able to buy them .\\
    \textbf{8} &  \textbf{i was glad i bought them .}\\
    9 &  i was able to buy a new pair .\\
    10 &  i was able to buy a new pair .\\
    11 &  i was able to buy a new pair of shoes .\\
    12 &  i was able to buy a new pair of new ones .\\
    13 &  i was able to buy a new pair of rubber flavored items .\\
    14 &  i was able to buy a new pair and i was very happy .\\
    \hline 
\end{tabular}
\begin{tabular}{|p{0.24\columnwidth}p{0.7\columnwidth}|}
    \hline
    \multicolumn{2}{|l|}{\textbf{Verbal Predicates}} \\
    \textbf{wish, known}  & \textbf{i wish i will always recognize them .} \\
    got & i got a new one .\\
    decided & i never decided on the new restaurants .\\
    went & i went home with a new friend .\\
    is & it is a great price .\\
    get & now i have to get a new one .\\
    felt & i felt very disappointed .\\
    go & i will go to the grocery store .\\
    took & i took them to the store .\\
    left & i left the store with a new tip .\\
    realized & after many years , i realized that .\\
    loved & i loved them .\\
    ran & i ran back to the store .\\
    \hline 
\end{tabular}
\begin{tabular}{|p{0.36\columnwidth}p{0.58\columnwidth}|}
    \hline
    \textbf{Frame Semantics} & \\
    \textbf{gold frame set*}
  & \textbf{i wanted to be a professional photographer .} \\
    Arriving & i got the earbuds .\\
    Quantity & it was a lot of fun .\\
    Becoming & it ended up being a target .\\
    Cardinal\_numbers & i paid for \$ 50 dollars .\\
    Being\_obligated & i guess i had a similar card .\\
    Kinship & my parents were proud .\\
    Statement & i told them i would not be a target .\\
    Causation & they sent me to the seller 's desk .\\
    Opinion & i think i was a millionaire . \\
    Perception\_experience & it was a hit . \\
    \hline
\end{tabular}
\begin{tabular}{|p{0.18\columnwidth}p{0.76\columnwidth}|}
    \hline
    \multicolumn{2}{|l|}{\textbf{Clusters}} \\
    0 & i ended up buying a new one . \\
    1 & i bought a new pair of shoes . \\
    \textbf{2} & \textbf{i was a good price . } \\
    3 & i bought a new pair of shoes for free .\\
    4 & i decided to buy a new pair of headphones . \\ 
    \hline 
\end{tabular}
\begin{tabular}{|p{0.18\columnwidth}p{0.76\columnwidth}|}
    \hline
    \multicolumn{2}{|l|}{\textbf{Oracle BOW}} \\
    \hline
    - & i then wish that i had no time . \\
    \hline 
\end{tabular}
\caption{Generated continuations from our framework with different control attribute values. Boldface indicates attribute values of the human-written continuation. 
* = the frame set in the human-written continuation contains the following frames: \emph{Desiring}, \emph{Experiencer}, \emph{Event}, \emph{Being\_named}, and \emph{Entity}. 
} 
\label{tab:system-outputs-all}
\end{table}

In this section, we describe the control attributes we explored using our framework. Table~\ref{tab:system-outputs-all} shows examples of generated continuations for a single story context with several values for the control attributes described below. Given our simple modeling framework, it would be natural to experiment with combining control attributes via summation or averaging of the attribute representations, but we leave an investigation of this to future work, focusing here on using one attribute at a time. 

\paragraph{Sentiment.} 
Stories may express sentiment regarding their characters or circumstances. We acquire sentiment labels by running the pretrained analyzer from \citet{D13-1170} on the continuations in the training data. The analyzer produces three labels: ``negative'', ``neutral'', or ``positive''. 

During training, $z_{\enc}$ and $z_{\dec}$ are fixed one-hot vectors for each value. 

\paragraph{Length.} 
Some prior work has generated summaries with a desired length~\citep{kikuchi-EtAl:2016:EMNLP2016,W18-2706}. We similarly use length of the continuation as a control attribute. Instead of using an embedding for each integer length value, we group the lengths into a small number of bins (details are provided below). $z_{\enc}$ and $z_{\dec}$ are fixed one-hot vectors for each bin.

\paragraph{Verbal Predicates.}  
Semantic role labeling (SRL) is a form of shallow semantic parsing that annotates predicates and their arguments in sentences. We consider predicates from a semantic role labeling as control attributes. We use the SRL system from AllenNLP~\citep{Gardner2017AllenNLP} to automatically obtain predicates for the continuations in our training set. Then, a predicate vector is obtained by first summing up 100-dimensional GloVe embeddings~\citep{pennington2014glove} of the predicted predicates (if there is more than one), then reducing the dimension to 64 using principal component analysis.\footnote{The reason we use PCA here is to make all attribute embeddings have comparable embedding size, though we did not systematically evaluate the effect of this choice.} We wish to clarify that we do not use the argument structure from the SRL system. We restrict our focus to simply the set of verbal predicates in the SRL structure; this would presumably be simpler to use in interactive settings where users would specify attribute values for generating continuations.

\paragraph{Frame Semantics.} 
A story is composed of a sequence of meaningful events~\citep{story}, often following particular patterns described in various terms such as scripts~\citep{schank+abelson77} and frames. FrameNet~\citep{BakerFillmoreLowe:98} is an inventory of semantic frames, which are semantic abstractions describing universal categories of events, concepts, and relationships.  

We consider frame semantics as another control attribute in our framework. In order to get a frame semantic representation for a continuation, we use SEMAFOR~\citep{frameparsing}. SEMAFOR automatically produces a frame-semantic parse for a sentence, which consists of spans that evoke particular frames in FrameNet as well as annotations of textual spans that correspond to frame-specific arguments. For our purposes, we drop the arguments and only use the set containing all frames that are evoked in the sentence. A sentence may contain multiple frames. For example, in the sentence ``Roa's advice made Emma a lot happier in her life!'', ``a lot'' evokes the \emph{Quantity} frame while ``Emma a lot happier'' evokes the \emph{Effect} frame.

The frame set variable $z$ is computed by summing embeddings for the frames in the set:
\begin{equation}
z = R(l)=\sum_{j \in l}R_j \label{table:sum_vec}
\end{equation}
where $l$ is the frame set and $R_j$ is the representation of frame $j$. The frame embeddings are learned during training.\footnote{The dimension of the frame vector is 64 in our experiments.} For modeling purposes, we restrict our attention to the 100 most frequent frames in the training data. The rest of the frames are pooled together to form a single additional ``catch-all'' frame.

\paragraph{Automatically-Induced Clusters.} We also experiment with running $k$-means clustering on the bag-of-words sentence representations of the continuations in the training set. 
We treat these automatically-induced cluster labels as control attribute values. Below we describe experiments with different cluster labels and analyze the characteristics of the generated outputs.

\paragraph{Oracle Bag-of-Words Sentence Representations.} 
We also consider the use of a bag-of-words (BOW) sentence representation as a control attribute. Naturally, the sentence representation of the continuation is not available before generating the continuation in practice. However, we can use this attribute to verify the capability of our model to reconstruct the continuation from its bag-of-words representation.

\section{Experimental Setup}

\subsection{Datasets} 

We experiment with the publicly available ROC story corpus developed by \citet{mostafazadeh-EtAl:2016}. 
It consists of approximately 100K five-sentence stories of everyday events. We sample 2000 stories as a development set and 2000 as our test set. The remaining stories form our training set. Our goal is to generate the fifth sentence (the ``continuation'') given the previous four sentences. We use the 10k most frequent words in the training set as our vocabulary. A special token $\langle \mathit{unk}\rangle$ is introduced for unknown words. 

\subsection{Evaluation}
Previous work evaluates generation tasks with automatic metrics, such as perplexity (\textbf{PPL}), \textbf{BLEU}~\citep{bleu},\footnote{In this paper, all BLEU scores are BLEU-2 scores (i.e., using unigrams and bigrams).} 

and \textbf{ROUGE}~\citep{Lin04rouge}. 
We adopt these in our evaluation and add three more metrics using the pretrained story scorer from \citet{sagarkar-18}. The scorer rates a generated continuation given its context along three dimensions: \textbf{relevance (R)}, \textbf{interestingness (I)}, and \textbf{overall quality (O)}. The story scorer does not use a gold standard continuation.

In addition, to evaluate the diversity of the generation, we use \textbf{Max-BLEU}\footnote{While max metric scores are not solely measuring diversity, they do provide a sense of the potential of the list. If all entries on the list are the same, the max metric scores would equal the average metric scores. The difference between the max and average metric scores therefore can be viewed as providing a bound on the diversity.} and \textbf{Max-ROUGE}. 
 
First, we compute BLEU and ROUGE scores over a set of outputs $(y_1, y_2, ..., y_n)$ with different attribute values given the same story context, then we compute the max scores: 
\begin{align*}
\maxbleu &= \max_{i} \bleu (y_i, r) \\
\maxrouge &= \max_{i} \rouge (y_i, r) 
\end{align*}
where $r$ is the gold standard continuation. 

We also use \textbf{Self-BLEU}~\citep{zhu2018texygen} to evaluate the diversity of a set of outputs. It is calculated by averaging the BLEU scores computed between all pairs of generated continuations for a given context, then averaging this quantity over all contexts. 

The smaller the Self-BLEU score is, the more diverse are the generated outputs.

\subsection{Training Details}
Our seq2seq model has a 2-layer biLSTM~\citep{Hochreiter:1997:LSM:1246443.1246450} encoder and a 1-layer LSTM decoder. The hidden dimension of all layers is 512. The word embedding dimension is also 512. For optimization, we use Adam~\cite{kingma2014adam} with learning rate 0.001. We use early stopping based on perplexity on the development set.

\section{Results}
We now present our experimental results. Section~\ref{sec:controllability} includes results related to how well our generated output matches the desired attribute values. Section~\ref{sec:oracle} presents results when generating continuations with oracle attribute values. 
In Section~\ref{sec:evalsets} we use our set-level metrics to evaluate sets of outputs with various attribute values. In Section~\ref{automic_generation} we report results when attempting to automatically infer attribute values to generate a small set of high-quality outputs. 

\subsection{Controllability Evaluation}
\label{sec:controllability}
In this section, we evaluate the controllability accuracy of our framework by automatically measuring the match between the attribute values of the generated continuations and the desired values. 
For certain control variables, like sentiment and frames, this automatic evaluation is prone to errors in the associated analyzers. 
That is, the metrics that rely on automatic analyzers could become artificially high if our generation models learn to produce outputs that match the biases of the analyzers.
We could instead consider manual evaluation of control accuracy. However, we were more interested in devoting our manual evaluation to the question of whether users would find the system outputs useful for a particular goal.

\begin{table}[t]
\centering
\small
\begin{tabular}{l|c|c|c|}
\cline{2-4}
& \multicolumn{3}{c|}{\bf{Generated Continuations}}
\\ \cline{2-4} {\bf{Target Sentiment}}
& negative & neutral & positive  \\\hline
\multicolumn{1}{|l|}{negative} & \bf{68.5} & 19.6  & 12.0  \\ 
\multicolumn{1}{|l|}{neutral} & \phantom{0}7.0 & \bf{73.9}  & 19.2    \\ 
\multicolumn{1}{|l|}{positive} & \phantom{0}0.8 & \phantom{0}3.1 & \bf{96.2}  \\ \hline
\end{tabular}
\caption{Sentiment match percentages of generated continuations and target sentiment values.}
\label{table:sentiment_analysis}
\end{table}

\begin{table}[t]
\centering
\small
\begin{tabular}{c|c|c|c|c|}
\cline{2-5}
& $\dif=0$ & $ \dif \leq 1$  & $ \dif \leq 2$ & $\dif \leq 3$ \\ \hline
\multicolumn{1}{|c|}{3 bins} & 95.8  & 100 & - & -\\
\multicolumn{1}{|c|}{30 bins} & 70.35  & 94.8 & 99.25 & 99.9 \\ \hline
\end{tabular}
\caption{Frequency (\%) of the generated continuations in the range of $\dif=|l -l_p|$ where $l$ is the continuation length and $l_p$ is the desired length.}
\label{table:length_result}
\end{table}

\paragraph{Sentiment.}

We generate three continuations for each story in the development set, one for each sentiment label. Using the same sentiment analyzer from \citet{D13-1170} as above, 
we obtain predicted sentiment labels for the continuations. Table \ref{table:sentiment_analysis} shows the sentiment distribution for each label. We see that the vast majority of the time, the continuations match the desired values. Matching positive sentiment is easiest for our model, followed by neutral. 

\paragraph{Length.}

We quantize the generation lengths into bins, each representing a size range. Below are the two settings we consider:
\begin{itemize}
\item 3 bins: We use three bins with the following length ranges: 
[1,7], 
[8,13], and 
[14, $\infty$). 

\item 30 bins: We use a bin for each length. No sentence is longer than 30. 
\end{itemize}

During training, we do not update the representations of the length control variable. After training, we treat the length of the continuation in the development set as the target control variable and generate continuations for each length. The results are shown in Table~\ref{table:length_result} and demonstrate that our model can generate continuations with the desired length with only small differences.

\begin{table}[t]
\centering
\small
\begin{tabular}{|l|r|l|r|}
\hline \bf Predicate & \bf M\% & \bf Predicate &   \bf M\%  \\ \hline
was &  100 & got & 100 \\ 
had & 100 & decided & 94.4 \\
went & 99.9 & is & 100 \\
 made  & 99.25 & were & 100 \\
 found & 100& get & 99.95\\
felt &  99.55 & go & 100 \\
took & 99.2& ended & 98.25 \\
be  & 99.95 & told & 99.9  \\
gave   & 99.95 & left & 99.85 \\
said & 100& bought & 100\\
\hline
\end{tabular}
\caption{\label{srl} Match percentages (M\%) showing fraction of stories for which generated continuations contain the desired predicate. 
The 20 most frequent predicates are shown; additional results are in the Appendix.}
\end{table}

\paragraph{Verbal Predicates.}

We select the top 100 most frequent verbal predicates in the training data. Then for all the stories in the development set, we generate a continuation for each of the 100 predicates. We check whether the predicate appears in the generated continuations. As the results in Table \ref{srl} show, the framework can nearly always generate outputs with the desired predicates.

\begin{table}[t]
\setlength{\tabcolsep}{3pt}
\centering
\small
\begin{tabular}{|l|l|l|l|}
\hline \bf Frame & \bf M\% & \bf Frame &   \bf M\%   \\ \hline
Calendric\_unit & 99.5& Locative\_relation & 87.4 \\ 
 Arriving & 98.9 & Quantity & 92.5 \\
Temporal\_collocation & 99.9 & Becoming& 99.6 \\
Cardinal\_numbers& 89.1 &Being\_obligated & 97.7 \\
Kinship& 94.4& Intentionally\_act& 99.7 \\
Statement& 98.0 & Causation& 98.6 \\
Emotion\_directed & 98.7& Buildings& 93.0 \\
Personal\_relationship& 92.7 & Food& 79.2 \\
Self\_motion& 86.4&Capability & 99.9 \\
Desirability & 98.1& Observable\_body\_parts& 74.2\\
\hline
\end{tabular}
\caption{\label{frame} Match percentages (M\%) showing fraction of stories for which generated continuations contain the desired frame. Additional results are in the Appendix.}
\end{table}

\begin{table}[t]
\centering
\small
\begin{tabular}{c|c|c|c|c|c|}
\cline{2-6}
& \multicolumn{5}{c|}{\bf{Generated Continuations}}
\\ \cline{2-6} {\bf{Target Value}}
& 0 & 1 & 2 & 3 & 4  \\\hline
\multicolumn{1}{|c|}{0} & \bf{79.9} & \phantom{0}2.8  & \phantom{0}3.1  & \phantom{0}0.9 & 13.4\\ 
\multicolumn{1}{|c|}{1} & \phantom{0}5.1 & \bf{63.1}  & 26.4 & \phantom{0}1.4 & \phantom{0}4.2  \\ 
\multicolumn{1}{|c|}{2} & \phantom{0}2.6 & \phantom{0}2.0 & \bf{90.6} & \phantom{0}0.3 & \phantom{0}4.7\\
\multicolumn{1}{|c|}{3} & 20.9 & 20.1 & 24.6 & \bf{31.0} & \phantom{0}3.5\\
\multicolumn{1}{|c|}{4} & \phantom{0}0.9 & \phantom{0}0.3 & \phantom{0}0.5 & \phantom{0}0.1 & \bf{98.3}\\ \hline
\end{tabular}
\caption{Cluster match percentages (\%) for each value of the cluster control variable.
}
\label{table:knn_analysis}
\end{table}

\begin{table*}[ht]
\centering
\small
\begin{tabular}{l|c||c|c||c||c|c|c|}
\cline{2-8} 
& \multicolumn{1}{c||}{\multirow{2}{*}{PPL ($\downarrow$)}}
& \multicolumn{2}{c||}{ROUGE}
& \multicolumn{1}{c||}{\multirow{2}{*}{BLEU ($\uparrow$)}}
& \multicolumn{3}{c|}{Story Scorer}
\\ 

& & \multicolumn{1}{c}{ROUGE-1 ($\uparrow$)} & ROUGE-L ($\uparrow$) & & \multicolumn{1}{c}{O ($\uparrow$)} & \multicolumn{1}{c}{R ($\uparrow$)}  & I ($\uparrow$)  \\ \hline
\multicolumn{1}{|l|}{seq2seq} & 25.8  & 27.0 
& 23.5 & 17.7 & 5.5 & 5.5 & 4.8\\
\hline 
\multicolumn{1}{|l|}{sentiment} & 25.0 &  26.7 
& 23.5 & 17.7  & 5.5  & 5.6 & 4.8  \\
\multicolumn{1}{|l|}{length} & 23.1 & 27.3  
& 24.6 & 20.3 & 5.7 & 5.8 & 5.0\\ 
\multicolumn{1}{|l|}{predicates} & 17.1 &  42.9 
& 35.1 & 26.4  & 6.0  & 6.2 & 5.2\\
\multicolumn{1}{|l|}{frames} &  15.0 & 41.1 
& 35.0  & 27.2 & 5.9  & 6.1 & 5.2 \\  
\multicolumn{1}{|l|}{clusters} & 24.3  & 28.6 
& 25.0 & 18.4  & 5.5 & 6.1 & 5.1 \\
\hline
\multicolumn{1}{|l|}{BOW} & \phantom{0}5.7 & 64.5  
& 54.5 & 45.4 & 6.2 & 6.2 & 5.2\\

 \hline
\multicolumn{1}{|l|}{gold standard} &  - & -   & - & - &6.5  & 6.7 & 5.7  \\
\hline

\end{tabular}
\caption{Automatic metrics for baseline system and when using oracle values for control attributes. For the gold standard continuation, we report only the story scorer results because they do not require a gold standard (unlike the other metrics).}
\label{table:oracle_result}
\end{table*}

\paragraph{Frame Semantics.}

In order to check how frequently the generated output matches the desired frames, we generate continuations for the top 100 frames (one frame for each continuation) for all stories in the development set. We check whether the frame appears in the specific continuation using SEMAFOR. The results are shown in Table \ref{frame}. Most frames have very high match accuracies, but there are a few frames with much lower accuracy, such as ``Food'' and ``Observable\_body\_parts''. These are more concrete frames that may be difficult to reasonably incorporate in certain story contexts.

\paragraph{Automatically-Induced Clusters.}

Given the cluster, the model generates a continuation. Then, we represent the continuation as a bag-of-words sentence embedding (using the same method as when performing the initial clustering) and find the cluster that has the nearest cluster embedding. Then we check whether the two clusters match.

In analyzing the clusters, we observed that cluster 0 corresponds to simple but reasonable continuations. Cluster 2 corresponds to continuations with positive sentiment. Cluster 4 contains continuations with more actions.  Some of the generated outputs are shown in Table \ref{tab:system-outputs-all}. 
From the results in Table \ref{table:knn_analysis}, we still see controllability for most clusters; however, for target cluster 3, which is rather generic based on our observations, the generated output seems flat.
 
\subsection{Evaluation with Oracle Attributes}
\label{sec:oracle}

Table \ref{table:oracle_result} shows automatic metric scores with 
oracle attribute values, i.e., using the attribute values of the gold standard continuations. Unsurprisingly, compared with the seq2seq baseline, the perplexity decreases and the ROUGE and BLEU scores increase with oracle attributes.
We also find that the scores from the story scorer, which does not use the gold standard while scoring, also show improvements over the baseline. 
We note that frame semantics form one of the most useful control attributes, aside from those that use words directly. 

The oracle BOW representation of the gold standard continuation yields the lowest perplexity and highest ROUGE and BLEU scores. It is not surprising that using this attribute 
would be useful according to metrics that favor matching the gold standard. However, these results do show that our simple modeling framework can make use of the information in the control variable with a high degree of  effectiveness. In addition, while the scores from the story scorer are generally higher than for other control attributes, they are roughly on par with those when using predicates and frames.

\begin{table*}[t]
\centering
\small
\begin{tabular}{|rl|cc|cc||cc||c|}
\cline{3-9}
\multicolumn{1}{l}{} & 
& \multicolumn{4}{c||}{ROUGE}
& \multicolumn{2}{c||}{\multirow{2}{*}{BLEU ($\uparrow$)}}
& 
\\  
\multicolumn{1}{l}{} &
& \multicolumn{2}{c}{ROUGE-1 ($\uparrow$)} & \multicolumn{2}{c||}{ROUGE-L ($\uparrow$)} & & & Self-BLEU ($\downarrow$) \\ 
\multicolumn{1}{l}{}  && Max & \multicolumn{1}{c}{(Avg)} & Max &(Avg) & Max &(Avg) &  \\ \hline
 & \multicolumn{1}{l|}{BS, beam = 3} & 31.8 & \bf (26.7)  
& 27.3 & \bf (22.6)  & 19.7 & (17.0) & 50.5\\
& \multicolumn{1}{l|}{TS, $\tau = 0.5$} &32.5 & (25.5)  
& 27.8 & (21.6)  &20.3 & (16.3)  & 
27.0\\
\multirow{2}{*}{3 continuations:} &\multicolumn{1}{l|}{TS, $\tau = 0.6$} & 30.5 & (22.2) 
& 25.9 & (18.8)  & 19.0 & (14.8) & \bf 23.8 \\
& \multicolumn{1}{l|}{sentiment} &\bf 32.8 & (25.6)  
& \bf 28.9 & (22.5)   &  \bf 21.1 & \bf (17.1)& 
30.7\\ 
&\multicolumn{1}{l|}{predicted frames} & 30.8 & (22.5) 
&27.0 & (19.9)  & 19.7 & (15.7) & 30.3 \\
&\multicolumn{1}{l|}{frames + reranking} & 30.7 & (21.9) 
& 26.3 & (18.9)  & 18.8 & (14.7) & 25.8 \\
\hline

&\multicolumn{1}{l|}{BS, beam = 5} & 33.9 & \bf (26.3) 
& 29.4 & \bf (22.3)  &21.3 & (16.2)  & 
68.1\\
5 continuations: &\multicolumn{1}{l|}{TS, $\tau = 0.5$} &35.0 & (25.3) 
&  30.0 & (21.5) &21.7 & (16.2)  & \bf 40.8 \\
&\multicolumn{1}{l|}{clusters} & \bf 36.1 & (24.5) 
& \bf 31.7 & (21.4)  & \bf 22.9 & \bf (16.4)  & 43.8 \\ \hline

&\multicolumn{1}{l|}{BS, beam = 30} & 40.0 & \bf (25.4)  
&  34.3 & (20.8) & 25.6 & (16.0)  & 92.5\\
30 continuations: &\multicolumn{1}{l|}{TS, $\tau = 0.5$} & \bf 43.0 & \bf (25.4) 
& \bf 37.5 & \bf (21.6)   & \bf 28.1 & \bf (16.3)  & \bf 74.0\\ 
&\multicolumn{1}{l|}{length} &42.1 & (24.7)  
&35.9 & (20.0)  & 26.2 & (14.8) &82.2\\ \hline

&\multicolumn{1}{l|}{BS, beam = 100} & 44.4 & (25.0)  
& 38.6 & (20.6)  & 29.2 & (15.9) & 96.2\\
\multirow{2}{*}{100 continuations:}&\multicolumn{1}{l|}{TS, $\tau$= 0.5} & 47.8 & \bf (25.4) 
& 42.3 & \bf (21.6)  & \bf 32.1 & \bf (16.3) & 85.6\\

&\multicolumn{1}{l|}{frames (individual)} & 47.0 & (24.0) 
&41.2 & (20.8)  &29.8 & (15.5)  & \bf 72.1 \\

&\multicolumn{1}{l|}{frames (sets)} & \bf 48.3 & (23.1)  
& \bf 42.7 & (20.1)  & 31.2 & (15.1) & 75.5\\
\hline
\end{tabular}
\caption{Metric scores to evaluate the potential of a list of continuations. We report the maximum and average metric scores over the continuations in each list to evaluate the quality of the lists, and self-BLEU to evaluate diversity. Best results for each metric and each number of outputs are in bold.} 
\label{table:diversity}
\end{table*}

\subsection{Evaluating Sets of Continuations}
\label{sec:evalsets}

We now evaluate \emph{sets} of continuations using our set-level metrics. 
Standard methods to generate sets of outputs include beam search (BS) and temperature-based sampling (TS), which we use as baselines. 

TS with temperature $\tau$ corresponds to transforming probabilities $p_i$ as follows: $\hat{p}_i \propto p_i^{\frac{1}{\tau}}$. 

A high temperature $\tau$ leads to high variety in generated samples, but also more noise, while lower temperatures lead to samples with less noise but also less diversity.

For each attribute, we generate continuations for each of its values, and compare to BS and TS systems with the same number of outputs. 

For example, for sentiment, we generate continuations for each of the 3 sentiment values and compare to BS and TS with 3 continuations. 

Results are shown in Table \ref{table:diversity}. BS shows the least diversity (as evidenced by its high self-BLEU scores). However, it generally yields high average ROUGE and BLEU scores. TS does very well in terms of diversity, and this diversity enables it to produce higher max scores than BS, but it has lower averages when using small numbers of continuations (3 or 5). 

Our sentiment- and cluster-controlled systems outperform TS in max metric scores and BS in diversity (self-BLEU). They also have the highest average BLEU scores, though the differences are small. With 30 continuations, TS with $\tau = 0.5$ performs best across all metrics; this number of continuations appears to be well-suited for temperature $0.5$. As we move to 100 continuations, we find that using our frame control variable leads to better diversity than TS, suggesting that the move to 100 samples has introduced some amount of repetition. By contrast, the 100 distinct frames and frame sets yield better diversity.

\subsection{Automatically Choosing Attribute Values}
\label{automic_generation}

Using our framework, we can generate continuations with any attribute values. However, if we are interested in generating a single continuation, we do not know the ideal attribute values to use. So, we propose two methods to automatically select a small set of values for the frame attribute.

\paragraph{Frames + Reranking:} Following \citet{LiGBGD15}, we rerank the outputs from the 100 most frequent frame sets by linearly combining the forward score $p(y\mid x)$ and the ``reverse score'' $ \lambda p(x\mid y)$, where the latter comes from a separately-trained seq2seq model. The forward score $p(y\mid x)$ is adjusted by dividing by the length of $y$ in order to not favor shorter outputs.

\paragraph{Predicted Frames:} We also build a model to automatically predict the frames in the continuation.
Given the frames in a sentence $x$, we compute a binary frame vector $f_x$ where 
entry $j$ is 1 if frame $j$ appears in $x$. We train a model that predicts the frame vector of the continuation given the frame vectors of the previous 4 sentences.  The model is an LSTM followed by averaging of hidden states. Mean squared error is minimized during training. After training, the $k$ continuations are selected based on the $k$ frames with the highest predicted score under this frame prediction model.

We use these two methods to produce 3 continuations for each story and report results in Table~\ref{table:diversity}. They both achieve a similar balance of quality and diversity as TS with $\tau=0.6$, with reranking leading to greater diversity than frame prediction and the latter showing higher ROUGE/BLEU scores.

\section{Human Evaluation}
Our previous results demonstrate that our frame control system has strong controllability and diversity in generation. In this section, we conduct a human evaluation to assess the utility of providing multiple suggestions from our models in a creative writing setting. We consider four different systems: \textbf{BS} with beam size 3; \textbf{TS} with 3 continuations using $\tau=0.6$, which we found to produce outputs with more diversity than $0.5$; \textbf{reranking} the 100 most frequent frame sets and using the top 3; and using continuations from the top-3 \textbf{predicted} frames under our frame prediction model.\footnote{The BS and TS baselines do not use control variables.}

To assess which set of generations from these four systems are most helpful in a collaborative writing setting, we collect annotations using Amazon Mechanical Turk. We randomly select 100 stories. For each story, we generate three outputs as a set of suggestions for each system, so there are 600 comparision pairs in total. We show workers two sets of outputs from different systems and ask them to select which suggestion is more helpful for writing the next line in the story. We also provide a choice of ``neither one is helpful at all''. We ask them explicitly to imagine they are writing the next line of the given story (see the appendix for more details).

Table \ref{table:human_analysis1} shows the results.\footnote{We remove results from 10-question sets where more than half of the questions were answered with the ``neither'' option, as we were concerned that these annotators did not fully understand the task or did not spend enough time studying the continuations. This occurred in roughly one third of question sets.} 
We observe that workers prefer the BS baseline over TS, although TS yields higher diversity. This could be because the continuations from BS are shorter, simpler, and more fluent. In addition, we observe that workers prefer the outputs from the reranking system over BS more often than not. Although the predicted frame system yields more diverse outputs, workers still prefer BS, likely due to the difficulty in predicting frames.
The reranking and predicted frame systems are both preferred to TS, though the gap is smaller with the predicted system. We also see that generating helpful suggestions is a difficult task: in many cases workers thought neither system was helpful, especially when given the outputs from BS/TS or TS/predicted.

One may ask why workers do not show a stronger preference for the more diverse sets of outputs. From our own preliminary annotations, we believe this is because diverse outputs tend to be longer and harder to understand, and also because greater diversity increases the chance of producing disfluent or nonsensical outputs. The BS outputs, by comparison, are sensical and mostly on-topic. Even if the suggestions are not creative, they may still help a worker to think about a new direction for the story to take. Nonsensical or disfluent suggestions, however, are rarely helpful. 

\begin{table}[t]
\centering
\small
\begin{tabular}{|cc|ccc|}
\cline{3-5}
\multicolumn{2}{c}{} & \multicolumn{3}{|c|}{\bf{Human Preference}}\\
\multicolumn{1}{c}{system 1} & system 2 
& 1 & 2  & neither \\ \hline
BS & TS
& 43 & 16 & 29\\ 
BS & reranking 
 
& 18 & 30 & 15\\ 
BS & predicted

& 38 & 29 & 19\\ 
TS & reranking

& 18 & 38 & 16\\
TS & predicted

& 18 & 27 & 34\\
reranking & predicted

& 27 & 24 & 21\\
\hline
\end{tabular}
\caption{Human preferences when given three continuations from each pair of systems.}
\label{table:human_analysis1}
\end{table}

\section{Related Work}

Automatic story generation has a long history, with early work based primarily on hand-written rules \citep{klein-73, meehan1977tale, dehn1981story, turner1993minstrel}. Subsequent methods were based on planning from artificial intelligence \citep{theune2003virtual, oinonen-2006, riedl2010narrative} and,  more recently, data-driven methods have been developed
\citep{mcintyre-lapata:2010:ACL,elson-12, daza-calvo-figueroanazuno:2016:CLfL2016,roemmele2015creative,N18-1204,martin2018event,fan-etal-2018-hierarchical,yao2019plan,fan-etal-2019-strategies}. In concurrent work, \citet{gupta-etal-2019-writerforcing} also propose methods to generate
more diverse and interesting story endings, albeit without control variables. 

In stronger relevance to our work, \citet{ClarkRTJS18} explore a creative writing setting with a machine in the loop, albeit with mixed results in terms of the quality of system suggestions. 
Predicting and controlling with frame values suggests a new way of interacting with collaborative writing systems, as long as frames can be communicated to users in ways they can easily understand. Recently, \citet{N18-1204} proposed a neural text generation method that explicitly represents and tracks entities. 
In addition, event sequences~\citep{D17-1168,DBLP:journals/corr/abs-1805-06122} are important elements in narrative texts but under-explored for story generation. These and related characteristics of creative writing could be incorporated into our framework as control attributes in future work.

The broader neural text generation community has also recently been interested in controllable text generation, i.e., generating text with specified characteristics reflected by control variables.  
In some previous work, the variables to be
controlled are embedded into vectors which are
then fed into models to reflect the characteristics
of the variables. \citet{kikuchi-EtAl:2016:EMNLP2016} and \citet{W18-2706} developed methods for controllable summarization, for example permitting users to control the length of the generated summary.
Related work has controlled style, topic, and sentiment polarity~\citep{DBLP:conf/icml/HuYLSX17,DBLP:journals/corr/abs-1709-03010,shen-1,DBLP:journals/corr/abs-1805-11749}. 

Despite the widespread usage of beam search for neural text generation, it has long been observed that its outputs are lacking in diversity. Several efforts have been made to provide diverse outputs for generation tasks, such as dialogue~\citep{LiGBGD15} and machine translation~\citep{N12-1059,gimpel-EtAl:2013:EMNLP,Li2016MutualIA}. Diverse beam search~\citep{VijayakumarCSSL16} produces a list of outputs with a diversity-augmented objective. \citet{ippolito-etal-2019-comparison} compare several methods for producing a set of diverse outputs from conditional language models. We leave a careful comparison to such algorithms to future work.

\section{Conclusion and Future Work}
We proposed a controllable framework that generates the next sentence of a story given its context. We experimented with a broad range of control attributes and demonstrated that our framework can accurately generate outputs that match the target values. 
Sets of outputs from our method show high diversity and high oracle metric scores. 
The human evaluation shows that the multiple suggestions from our model show promise for integration in a collaborative writing system. Future work could explore other control attributes as well as a compositional framework to control multiple attributes jointly.

\section*{Acknowledgments}
We would like to thank TTIC SL@TTIC researchers who participated in a preliminary annotation study. 

\bibliography{acl2019}
\bibliographystyle{acl_natbib}

\appendix

\section{Appendix}
\paragraph{Model.}

\begin{figure*}[t]
\centering
\includegraphics[width=0.8\textwidth]{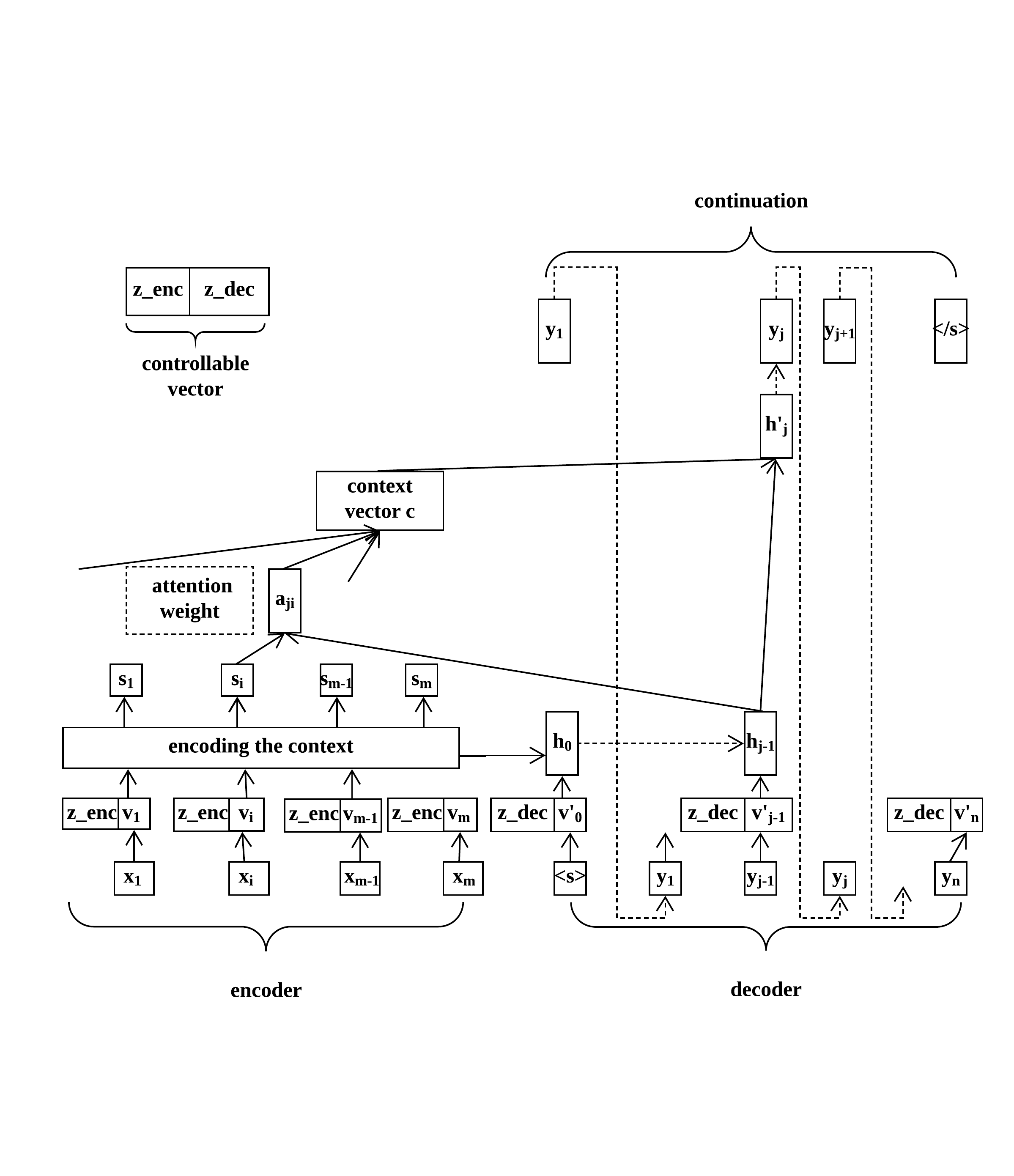}
\vspace{-1.4cm}
\caption{Our modeling framework for controllable generation. 
\label{fig:model}}
\end{figure*}

Figure \ref{fig:model} shows our modeling framework.

\paragraph{Additional Experimental Results.}

\begin{figure*}[h!]
\centering
\includegraphics[width=0.9\textwidth]{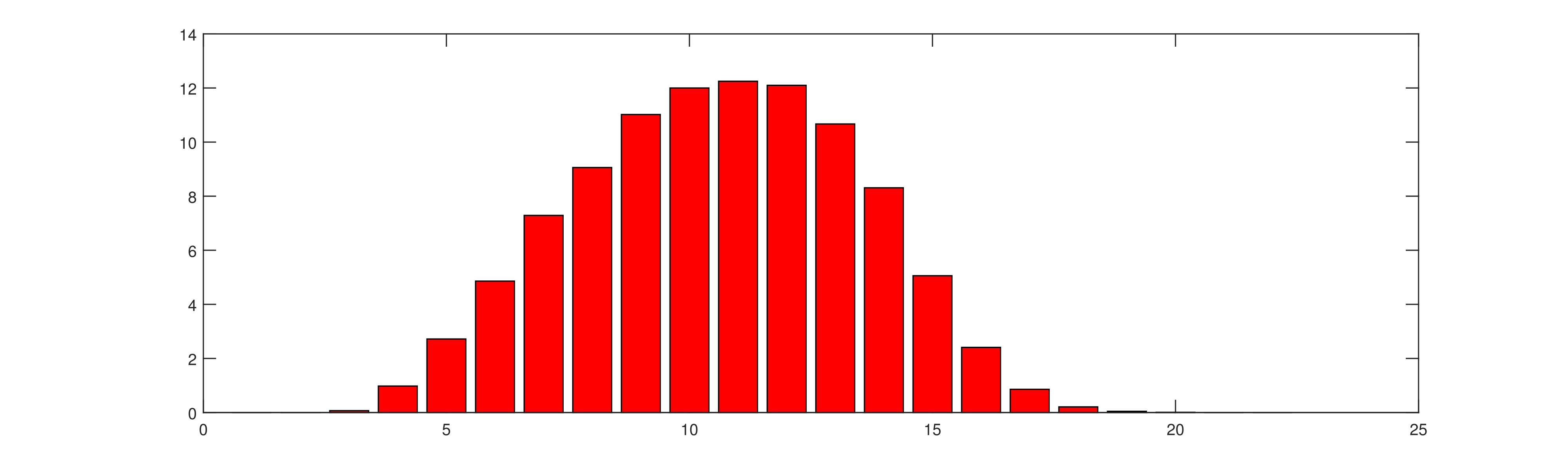}
\caption{The length distribution of the continuations in the training data. The horizontal axis shows continuation lengths and the vertical axis shows the frequency (\%) of the length.
\label{fig:length_data}}
\end{figure*}

\begin{figure*}[h!]
\centering
\includegraphics[width=0.5\textwidth]{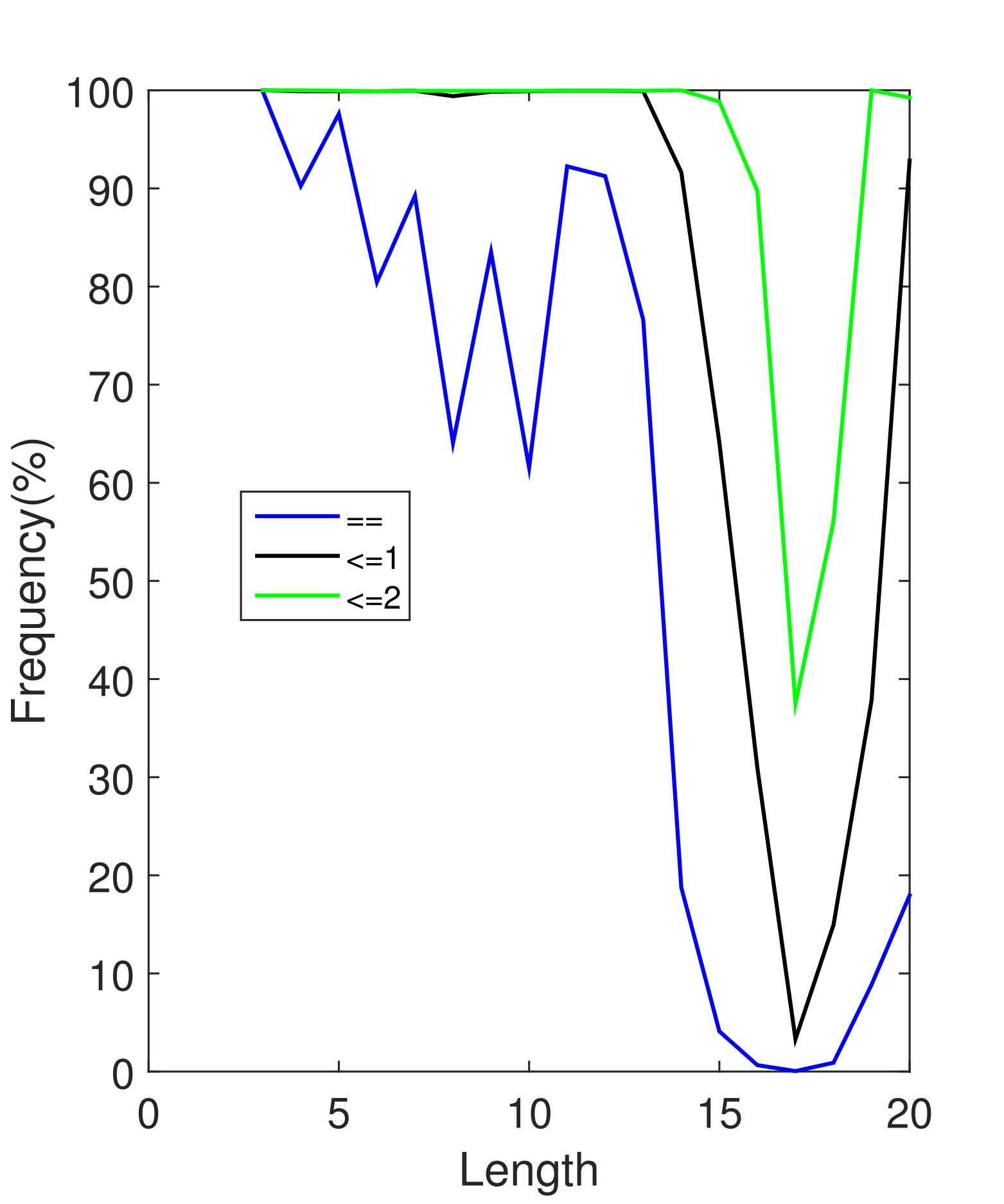}
\caption{Frequency (\%) of the generated continuations for different ranges of $\dif=|l -l_p|$. The horizontal axis is the desired length. The vertical axis is the percentage of generated outputs with the given values of $\dif$. 
Here $l$ is the story continuation length and $l_p$ is generated sentence length as the target control variable. ''$==$'' refers to $l=l_p$, ''$<=i$'' refers to $|l-l_p|<=i$.
\label{fig:length_cluster}}
\end{figure*}

Figure \ref{fig:length_data} shows the length distribution of the continuations in the training data. Figure \ref{fig:length_cluster}, Table \ref{srl_more}, and Table \ref{frame_more} show the frequency (\%)
of matching the target control variable for length, predicates, and frames, respectively.

\begin{table*}[h!]
\begin{center}
\small
\begin{tabular}{|r|r|r|r|r|r|r|r|r|r|}
\hline \bf predicate & \bf acc & \bf predicate &   \bf acc  & \bf predicate &   \bf acc & \bf predicate &   \bf acc & \bf predicate &   \bf acc \\ \hline
was &  100 & got & 100& had & 100& decided & 94.4 & went & 99.9\\
 is & 100 & made  & 99.25 & were & 100& found & 100& get & 99.95\\
felt &  99.55 & go & 100& took & 99.2& ended & 98.25& be  & 99.95\\
 told & 99.9  & gave   & 99.95 & left & 99.85 & said & 100& bought & 100\\
 came& 98.7 & realized & 79.75& loved  & 99.8& have & 99.1& won &100\\
 became &  99.75 & put & 83.6 & ran & 100 & see & 99.9&saw &98.75\\
 started & 99.85 & buy & 100& make & 97.3 & knew & 99.6& looked  &98.65\\
 lost & 99.05 & enjoyed &99.95 & called & 98.5 & learned & 99.8& ate  &95.35\\
 turned & 93.05& do & 100& wait  & 99.85 & take & 99.8&fell  &99.35\\
 wanted &  90.25& find &100 & thought & 80.05 & stopped & 98.65 & play & 100\\
 has & 98.15& eat & 100& let & 100& walked & 99.9& did & 96.55\\
 being & 99.05& done & 94.05& returned & 99.6& laughed  & 86.6& tried & 99.6\\
  's & 99.25& asked   & 99.9& began  & 93.75& helped   & 97.65& caught & 99.6\\
broke  & 99.9& getting & 99.95& keep & 97.15& paid  & 97.1& been  & 97\\
finished &99.9 & threw  &99.4 & spent & 99.85& passed  & 99.95& help &98.85 \\
are    & 99.65& drove & 99.25& work & 81.25& going  &35.9 & leave & 99.85\\
brought   & 98.4& feel & 99.75& picked & 73.65& agreed  & 98.35 & needed  & 24.2\\
worded    & 100 & used & 98.75& pay & 100& kept  & 98.8& liked & 98.65 \\
arrived & 99.2& played & 99.95& come & 98.7& thanked  & 98.3& relieved & 97.35\\
  stop  & 99.15& playing & 98.75& cried & 100& died  & 97.75& know & 89.05\\
\hline
\end{tabular}
\end{center}
\caption{\label{srl_more} Frequency (\%) of the generated continuations containing the desired predicate.}
\end{table*}

\begin{table*}[ht]
\setlength{\tabcolsep}{3pt}
\begin{center}
\small
\begin{tabular}{|l|l|l|l|l|l|l|l|}
\hline \bf frame & \bf acc & \bf frame &   \bf acc  & \bf frame &   \bf acc & \bf frame &   \bf acc \\ \hline
Calendric\_unit & 99.5& Locative\_relation & 87.4 & Arriving & 98.85 & Quantity & 92.5 \\
Temporal\_collocation & 99.9 & Becoming& 99.6 & Cardinal\_numbers& 89.05 &Being\_obligated & 97.65 \\
Kinship& 94.35& Intentionally\_act& 99.65 &
Statement& 98& Causation& 98.55 \\
Emotion\_directed & 98.65& Buildings& 93& Personal\_relationship& 92.7 &
Food& 79.2 \\
Self\_motion& 86.4&Capability & 99.95&Desirability & 98.1& Observable\_body\_parts& 74.15\\
Experiencer\_focus& 91.75&Time\_vector & 99& Request &99.35 &Deciding & 99.6  \\
Relational\_quantity & 95.7 &
Measure\_duration& 96.7& Frequency& 99.75& People& 90.8 \\
Ingestion& 86.25&Vehicle & 86.45 &Age& 98.55&Increment & 98.6 \\
Leadership& 80.6&Desiring & 90.7& Stimulus\_focus& 93.4 &
Activity\_start& 95.95 \\
Education\_teaching& 96.5&Degree & 100& Grasp& 92.7& Commerce\_buy & 98.7\\
Scrutiny& 93.5& Locale\_by\_use & 65.15& Conquering & 95.3& Giving & 99.3 \\
Clothing & 49.05 & 
Attempt& 98.95& Becoming\_aware & 90.25& Building\_subparts & 73.7 \\
Motion& 96.55& Placing& 73.15 &
Natural\_features& 26.7& Getting& 85.05 \\
Locating& 76.55&Sufficiency & 97.75&Feeling & 98.7 &
Awareness& 95.7 \\
Experiencer\_obj& 87.4& Working\_on & 67.65& Aggregate & 70.3&  Performers\_and\_roles& 87.95\\
Change\_position\_on\_a\_scale & 72.45& Roadways & 71.9& Containers & 44.1& Coming\_to\_believe & 92.4 \\
Assistance& 99.4 &
Ordinal\_numbers & 96.5& Relative\_time & 92.6& Choosing & 96.55  \\ Existence &88.15 & Dimension & 69.65 &
Cause\_harm & 89.75& Perception\_active & 87.1 \\
Text &63.6 & Cause\_motion & 81.2&  Possession& 76.85 &
Type& 78.2 \\
Body\_movement & 66& Opinion & 87.7& Removing & 74& Money & 97.5\\
Have\_as\_requirement & 25.4& Using & 93.1& Storing & 71.4& People\_by\_age & 57.25  \\
Contacting & 73.35 &
Make\_noise &97.9 & Substance & 50.4& Required\_event & 91.7  \\ Political\_locales & 85.3& Difficulty & 89.8 &
Activity\_ongoing & 94& Direction & 98.6  \\ 
Perception\_experience & 35& Impact &95.6 & Locale& 87.25 &
Waiting& 96.65 \\
Concessive & 95.15& Partitive & 63.65&Operate\_vehicle  & 61.55& Size & 98.3\\
\hline
\end{tabular}
\end{center}
\caption{\label{frame_more} Frequency (\%) of the generated continuations containing the desired frame.}
\end{table*}

\paragraph{Human Evaluation Setup.}

\begin{figure*}[t]
\centering
\includegraphics[width=0.95\textwidth]{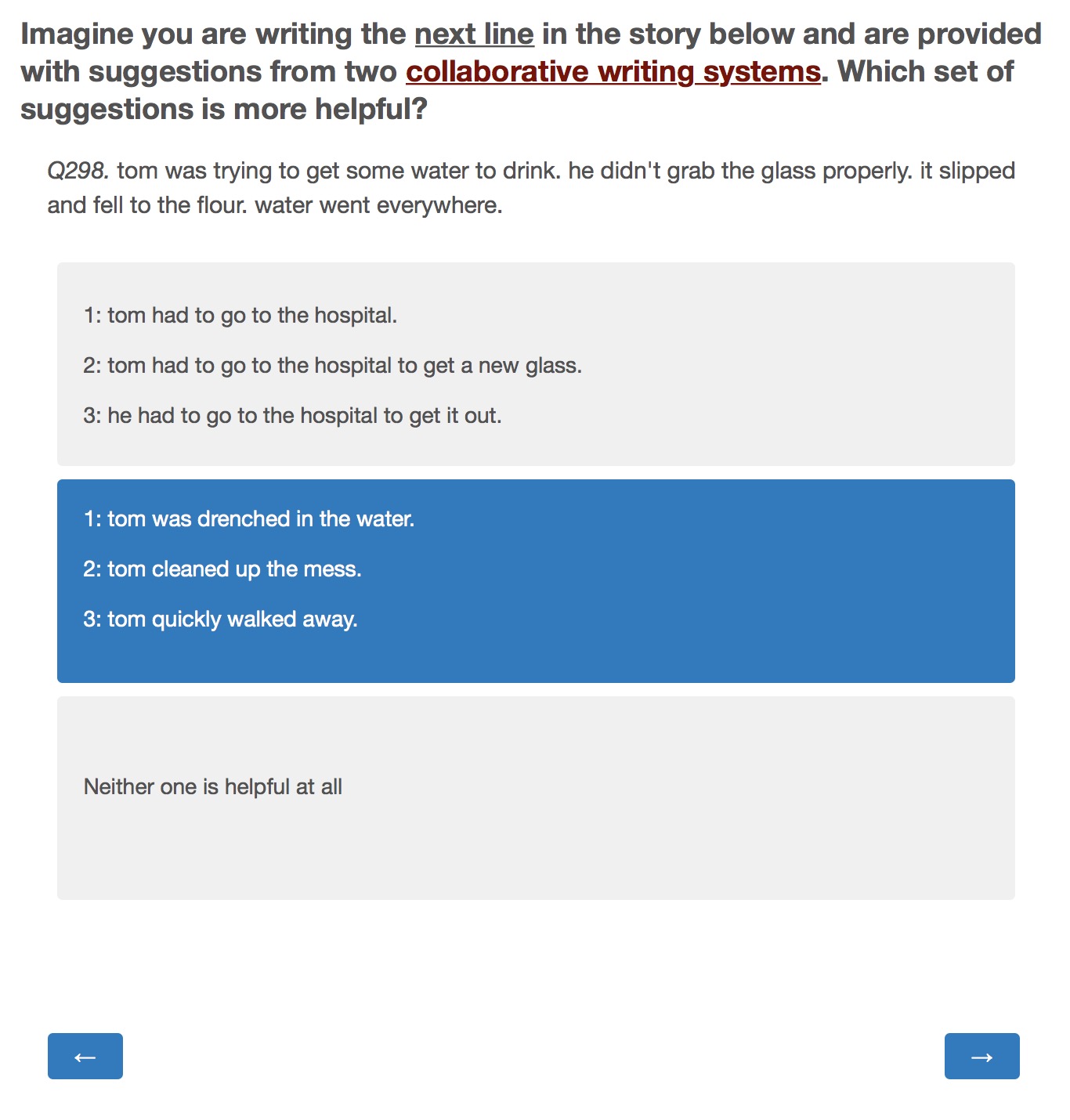}
\caption{The form for human evaluation of the generation systems for their potential in a collaborative writing setting.
\label{fig:tuker_eval}}
\end{figure*}
The form used on Amazon Mechanical Turk is shown in Figure \ref{fig:tuker_eval}.


\end{document}